\documentclass[runningheads]{llncs}
\usepackage{subfig}
\usepackage{graphicx}
\usepackage{amsmath,amssymb} 
\usepackage{color}
\usepackage{booktabs}
\usepackage{tabularx}
\usepackage{upquote}
\usepackage{multirow}
\usepackage{wrapfig}
\usepackage{caption}
\usepackage{bm}

\begin{document}
\title{Adaptive Future Frame Prediction with Ensemble Network}

\author{Wonjik Kim\inst{1,2}\orcidID{0000-0003-1067-8096} \and
Masayuki Tanaka\inst{1}\orcidID{0000-0003-4619-0033} \and
Masatoshi Okutomi\inst{1}\orcidID{0000-0001-5787-0742} \and Yoko Sasaki\inst{2}\orcidID{0000-0002-1220-6124}}
\authorrunning{W. Kim et al.}

\institute{Department of Systems and Control Engineering, School of Engineering, Tokyo Institute of Technology, Meguto-ku, Tokyo 152-8550, Japan
\email{wkim@ok.sc.e.titech.ac.jp; \{mtanaka, mxo\}@sc.e.titech.ac.jp}\\
\and
Artificial Intelligence Research Center, National Institute of Advanced Industrial Science and Technology, Koto-ku, Tokyo 135-0064, Japan\\
\email{y-sasaki@aist.go.jp}}
\maketitle              
\begin{abstract}
Future frame prediction in videos is a challenging problem because videos include complicated movements and large appearance changes. Learning-based future frame prediction approaches have been proposed in kinds of literature. A common limitation of the existing learning-based approaches is a mismatch of training data and test data. In the future frame prediction task, we can obtain the ground truth data by just waiting for a few frames. It means we can update the prediction model online in the test phase. Then, we propose an adaptive update framework for the future frame prediction task. The proposed adaptive updating framework consists of a pre-trained prediction network, a continuous-updating prediction network, and a weight estimation network. We also show that our pre-trained prediction model achieves comparable performance to the existing state-of-the-art approaches. We demonstrate that our approach outperforms existing methods especially for dynamically changing scenes.

\end{abstract}
\section{Introduction}
\label{sec:Intro}

\begin{figure}[t]
\begin{center}
  \includegraphics[width=100mm]{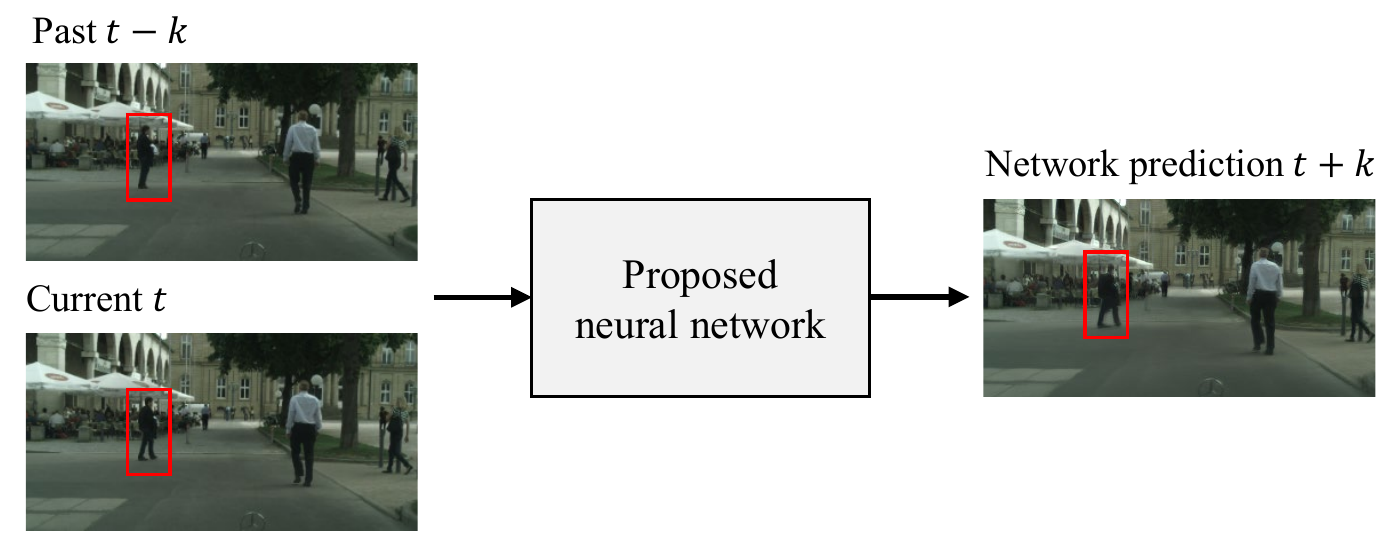}
  \caption{Future frame prediction of 2in-1out. The neural network gets two frames and predicts the future frame according to the inputs. In particular, we can confirm the prediction effects in the red-highlighted region.}
  \label{fig:fig1}
\end{center}
\end{figure}

Videos contain rich information, including movement and deformation of objects, occlusions, illumination changes, and camera movements. 
We can use that information for many computer vision applications.
Therefore, video analysis is actively researched to obtain such information~\cite{tekalp1995digital,bovik2009essential}. 
After the deep neural networks provided a positive impact on the computer vision in image domain~\cite{long2015fully,krizhevsky2012imagenet}, neural networks in the video domain are actively studied~\cite{liu2017video,lotter2016deep}. 
Future frame prediction described in Fig.~\ref{fig:fig1} is a branch of these research areas.

Video future frame prediction~\cite{wang2018predicting,pintea2014deja,NIPS2019_8637} is defined as estimating future frames from past frames. 
Video future frame prediction is a challenging task because videos include complicated appearance changes and complex motion dynamics of objects.
The uniform linear motion assumption does not work for the video future prediction. 
Then, learning-based deep network approaches have been researched~\cite{wang2018predicting,pintea2014deja,NIPS2019_8637}.
Existing future frame prediction algorithms based on the deep neural network can be classified in offline-supervised learning~\cite{wang2018predicting,pintea2014deja,NIPS2019_8637}.
In offline-supervised learning, a network is trained in advance with training data.
The limitation of offline-supervised learning is a mismatch of training data and real scenes.
The performance of the pre-trained network for new environment scenes which is not included in the training data is very low.
To overcome this limitation, we need to collect all possible scenes for the training data.
However, it is infeasible to collect all possible scenes.
One of the other approaches is adaptive online updating. 
If we can adaptively update the network model in the test phase, the prediction network can effectively work for new environment scenes.
It is a big advantage of adaptive online updating.
The problem of online updating is annotation. 
In the test phase, the annotation is required to obtain the ground truth. 
The annotation problem makes online updating difficult.
However, in the future frame prediction task, we can easily obtain the ground truth data, even in the test phase, by just waiting for a few frames. 
Then, in this paper, we propose an adaptive online update framework for the future frame prediction.
To the best of our knowledge, there is no existing research related to the adaptive online update framework for the future frame prediction.

The proposed adaptive online update framework consists of three sub-networks: a pre-trained prediction network, a continuous-updating prediction network, and a weight estimation network. 
The pre-trained prediction network is trained with the offline supervised learning manner as same as existing future frame prediction networks. The pre-trained prediction network model is fixed, while the continuous-updating prediction network model is updated in the test phase to adapt a huge varaety of scenes. The continuous-updating prediction network is relatively unstable because it tries to adapt to the current environment. Therefore, we also simultaneously train the weight estimation network online manner. Then, the final results are generated by ensembling the outputs of the pre-trained prediction network and the continuous-updating prediction network based on the estimated weight.

The main contributions of our work are:
\begin{itemize}
    \item[$-$] We propose an adaptive future frame prediction framework using network ensemble. 
    It is efficient for large-scale real scene prediction.
    \item[$-$] 
    We also propose the future prediction network model. 
    We demonstrate that the performance of this network is comparable to those of the existing state-of-the-art network models.
\end{itemize}

We have published the proposed network, data, and reproduction code on the website\footnote{http://www.ok.sc.e.titech.ac.jp/res/FFP/}.

\section{Related Works}
\label{sec:Related}
{\bf Online learning.}
In the real world, data distribution is continuously changing. An online learning algorithm has to include a mechanism for handling the various scenarios in a dynamic environment. This fact indicates that the parameters need to be updated automatically. In an evolving data environment, the adaption algorithm is essential to process proper decision making. 

According to~\cite{perez2018review}, there are three major online learning trends: sliding window, forgetting factor, and classifier ensembles. The sliding window~\cite{kuncheva2009window} is the most popular technique to handle the data flows. A window moves to include the latest data, then the model updates using the window. Therefore, the window size critically affects the model performance. The forgetting factor~\cite{ghazikhani2014online} incorporates a weight to maintain a balance in the streaming data. It classifies the data according to the importance and sets a weight. The importance can be determined with criteria such as age or relevance.

The other adaption algorithm is the classifier ensemble~\cite{minku2011ddd,ditzler2014domain}. The classifier ensemble combines several models to obtain a final solution. Several models include an adaptive model for storing the latest information for the current environment. One of the main advantages of this approach is that it is more likely to be more accurate than a single classifier due to reduced error distribution. Another advantage is that the ensemble method can efficiently deal with repeated known environments. To use the benefits, we employed a classifier ensemble concept for constructing the adaptive future frame prediction network.

\noindent
{\bf Future frame prediction.}
One of the popular approaches in future frame prediction is forecasting future frame pixel-values directly. Xingjian et al.~\cite{xingjian2015convolutional} used long-short-term-memory (LSTM) for frame prediction. Lotter et al.~\cite{lotter2016deep} built a predictive recurrent neural network to mimic predictive coding in neuroscience. However, these approaches suffer from blurriness in prediction and fail to extract large scale transformation. This can be attributed to the difficulty of generating pixel values directly, and low-resolution data sets with no significant movement~\cite{soomro2012ucf101,dollar2009pedestrian}. To address these problems, Wang et al.~\cite{wang2018predicting} focused on regions of interest to generate accurate predictions.

Another approach in future frame prediction is to estimate a future frame based on optical flows. 
Pintea et al.~\cite{pintea2014deja} assuming that similar motion is characterized by similar appearance, proposed a motion prediction method using optical flows. 
Liu et al.~\cite{liu2017video} designed a deep neural network that develops video frames from existing frames alternate to directly predicting pixel values. These approaches use optical flow estimation and warping function. Temporal consistency and spatial richness are guaranteed because the predicted pixels are derived from past frames. However, for the same reason, they fail if a place does not appear in past frames.

The other approach in future frame prediction is a multi-model composed of the above-mentioned approaches. Yan et al.~\cite{yan2019mixpred} studied future frame prediction with a multi-model composed of the temporal and spatial sub-network. Kim et al.~\cite{NIPS2019_8637} detected key-points in input frames, and translated the frame according to key-points motion. Gao et al.~\cite{Gao_2019_ICCV} initially predicted the optical flow and made warp a past frame with predicted optical flow. Subsequently, they refined the warped image in pixel-wise to generate a final prediction. This approach refines the image warped with forecast optical flow. This helps to manage places unseen in past frames.

Among the approaches mentioned above, the multi-model methods showed effectiveness in various scenes. Therefore, we designed the future frame prediction network as a multi-model structure. Because the future frame prediction network is a multi-model structure, it can be expected to handle the dynamic and various online-updating environment.

\section{Proposed Ensemble Network}
\label{sec:Proposed Online-Updating Prediction Network}
Future frame prediction forecasts future frames from past frames. We consider a prediction task of $2$in-$1$out that takes two frames ${\bm x}_{t}, {\bm x}_{t-k} \in \mathbb{R}^{W \times H \times 3}$ and predicts 
one future frame ${\bm x}_{t+k} \in \mathbb{R}^{W \times H \times 3}$
in the same interval, where $W$, $H$, and $3$ denote the width, height, and RGB channels of the frame respectively, and $t$ and $k$ are the current time, and the time interval respectively.

The entire process and network architecture are described in Section~\ref{subsec:Network Architecture}. In Section~\ref{subsec:Network Architecture}, the continuous-updating prediction network and weight estimation network which are online updating modules are also discussed in detail. Subsequently, the specification of the pre-trained prediction network which is a weight fixation module is explained in Section~\ref{subsec:Offline Network Architecture}. Finally, network training is described in Section~\ref{subsec:Network Training}.

\subsection{Overall Network Architecture}
\label{subsec:Network Architecture}

\begin{figure*}[t]
  \begin{center}
    \includegraphics[width=100mm]{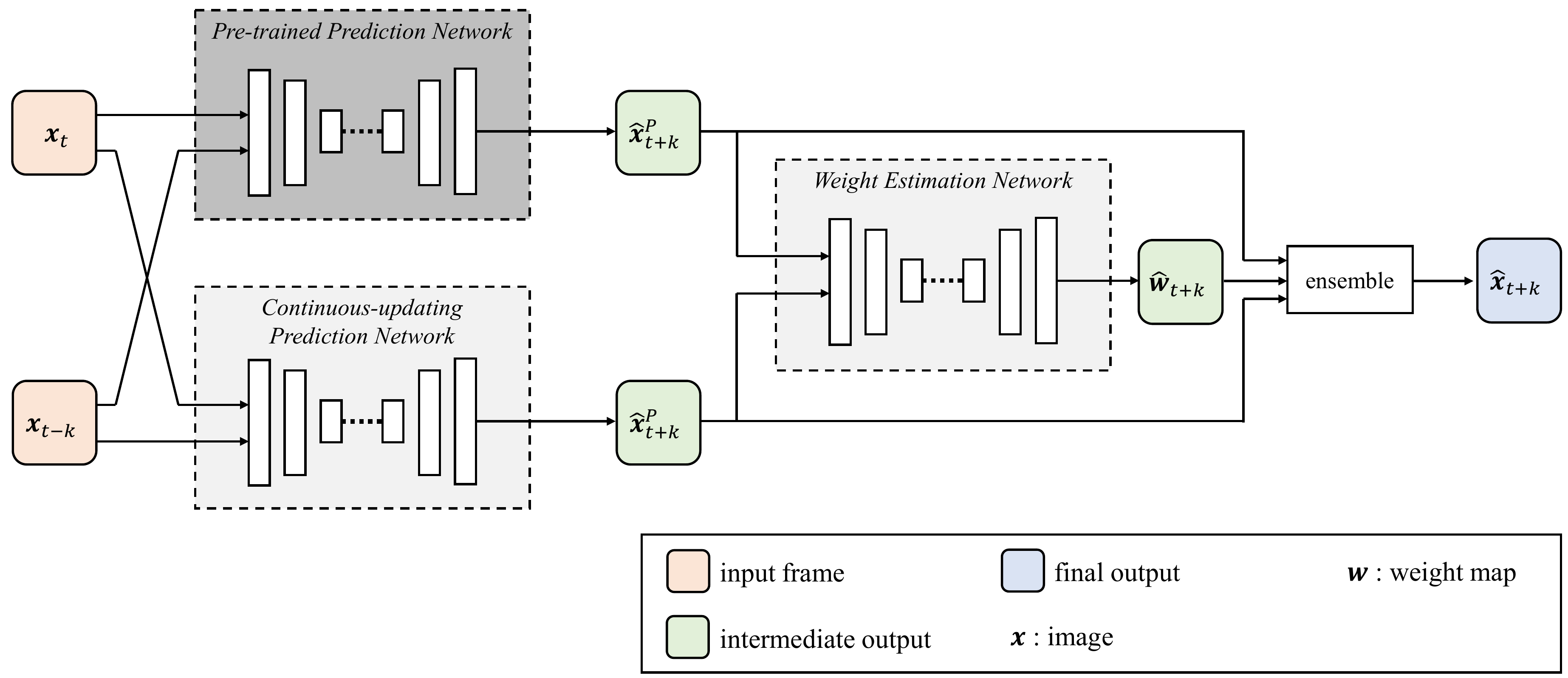}
  \caption{Our proposed ensemble network for future frame prediction. \label{fig:onlinearch}}
  \end{center}
\end{figure*}

As mentioned in Section 1, the pre-trained future prediction network only works for similar scenes of the training data. We can adaptively update the prediction network while the continuous-updating prediction network tends to be unstable because the continuous-updating prediction network tries to adapt to the current scene. Therefore, we propose an adaptive ensemble network which is combined with the pre-trained and the continuous-updating prediction networks.
The final prediction result is composed by combining the results of the pre-trained and the continuous-updating prediction networks based on the weight estimated by the weight estimation network as shown in Fig.~\ref{fig:onlinearch}. 

\subsubsection{Pre-trained Prediction Network}
\label{subsubsec:Fixed-Weight Network}
We train the pre-trained prediction network in advance with an offline learning manner.
This approach is well studied in the deep learning research area.
If test scenes are similar to the training data, the pre-trained prediction network can accurately predict the future frame. However, the pre-trained prediction network fails to predict untrained scenes. 
We use this pre-trained prediction network without updating weights in test phase.

\subsubsection{Continuous-updating Prediction Network}
\label{subsubsec:Adaptive-Weight Network}
First, we train the continuous-updating prediction network with an offline learning manner as same as the pre-trained prediction network. If the network structure of the continuous-updating prediction network is identical to that of the pre-trained prediction network, the wights of the continuous-updating prediction network can be initialized by the weights of the pre-trained prediction network.

The point of the continuous-updating prediction network is an adaptive updating process. 
At time $t$, we already have the frames ${\bm x}_{t-2k}, {\bm x}_{t-k}$, and ${\bm x}_{t}$. Then, we can update the weights of the continuous-updating prediction network, 
where the frames ${\bm x}_{t-2k}$ and ${\bm x}_{t-k}$ are used as inputs and the frame ${\bm x}_{t}$ can be used as the ground truth of the prediction.
The loss to update the continuous-updating prediction network at the time $t$ is
\begin{eqnarray}
 L_t^{C}({\bm \theta}_{C}) &=& \mu\left( {\bm x}_{t}, {\bm f}_{C}({\bm x}_{t-k}, {\bm x}_{t-2k}; {\bm \theta}_{C} ) \right) \,,
 \label{eq:LtA}
\end{eqnarray}
where $\mu$ is the loss function, ${\bm f}_{C}$ represents the prediction network, and ${\bm \theta}_{C}$ is the parameters of the network.
We call this updating process an online adaptive updating. 
We introduce an image quality measure $\mu$ for the loss function in Section~\ref{subsec:Network Training}.

\subsubsection{Weight Estimation Network}
\label{subsubsec:Weight Network}
In our framework, we have two prediction networks; the pre-trained prediction network and the continuous-updating prediction network. Those two networks are good for different situations. Then, we blend two outputs of the networks. For the blending, we train a weight estimation network to obtain the blending coefficients. 

\begin{eqnarray}
 \hat{\bm x}_{t+k} &=& \hat{\bm w}_{t+k} \otimes \hat{\bm x}_{t+k}^{P} 
 + (1-\hat{\bm w}_{t+k}) \otimes \hat{\bm x}_{t+k}^{C} \,,
 \label{eq:hatX}
\end{eqnarray}
where $\otimes$ represents elements-wise multiplication operator, 
and $\hat{\bm x}_{t+k}^{P}$, $\hat{\bm x}_{t+k}^{C}$, and $\hat{\bm w}_{t+k}$ are defined as
\begin{eqnarray}
 \hat{\bm x}_{t+k}^{P} &=& {\bm f}_{P}( {\bm x}_{t}, {\bm x}_{t-k}; {\bm \theta}_{P} ) \,, \\
 \hat{\bm x}_{t+k}^{C} &=& {\bm f}_{C}( {\bm x}_{t}, {\bm x}_{t-k}; {\bm \theta}_{C} ) \,, \\
 \hat{\bm w}_{t+k} &=& {\bm f}_{W}( {\bm x}_{t}, {\bm x}_{t-k}; {\bm \theta}_{W} ) \,.
\end{eqnarray}
Here, 
${\bm f}_{P}$ represents the pre-trained prediction network, 
${\bm \theta}_{P}$ is the parameters of the pre-trained prediction network,
${\bm f}_{W}$ represents the weight estimation network, 
and
${\bm \theta}_{W}$ is the parameters of the weight estimation network,

The pre-trained prediction network is fixed during the test phase, while the continuous-updating prediction network is updated in the test phase. 
Then, we need to train the weight estimation network by online updating.

In the early stage of the test phase or after a sudden scene change, the continuous-updating prediction network might be unstable. For those situations, the weight estimation network is trained to put a higher priority on the pre-trained prediction network. 
For the scene in which the training data does not include, the weight estimation network is trained to put a higher priority on the continuous-updating prediction network because the pre-trained prediction network fails to predict the scene.

\begin{figure*}[t]
  \begin{center}
    \includegraphics[width=100mm]{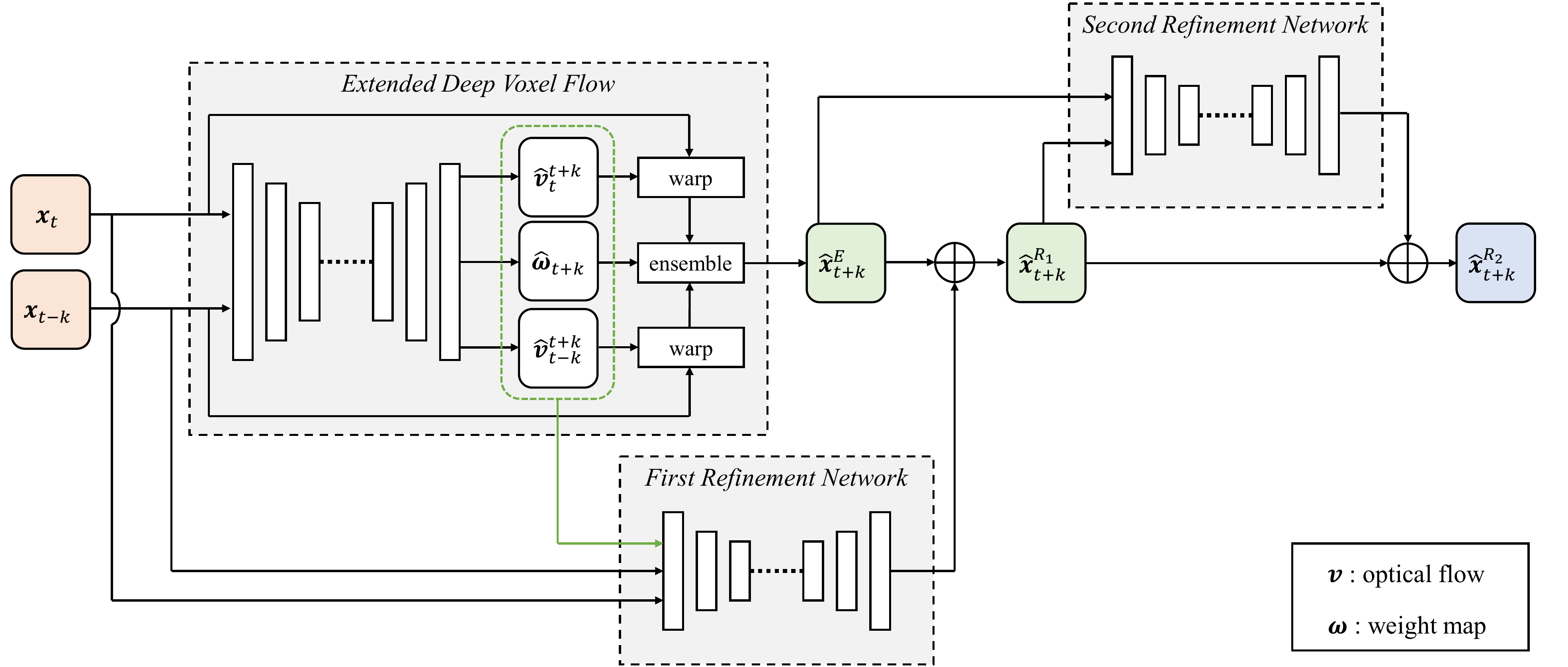}
  \caption{Future frame prediction network architecture. \label{fig:offline arch}}
  \end{center}
\end{figure*}

\subsection{Future Frame Prediction Network}
\label{subsec:Offline Network Architecture}

The proposed online future frame prediction framework internally includes the future frame prediction network as shown in Fig.~\ref{fig:onlinearch}.
We can independently design the network architectures for the pre-train prediction network and the continuous-updating prediction network. 
In this paper, we adopt to use the same network architecture for those two prediction networks for simplification of the network design. In addition, we can initialize the weight of the continuous-updating prediction network with those of the pre-trained prediction network if the network architectures of those two networks are the same.
In this section, we describe common network architecture for those two prediction networks.

We design the prediction network architecture inspired by~\cite{liu2017video,Gao_2019_ICCV}.
Our prediction network is composed of an extended deep voxel flow (EDVF) network with two refinement networks.
The EDVF is an extended version of DVF~\cite{liu2017video}. 
In the original DVF, they assume that the optical flow from $t-k$ frame to $t+k$ frame is just twice the optical flow from $t$ frame to $t+k$ frame. 
It means that they implicitly assume a linear uniform motion for the optical flow. 
In our EDVF, we separately estimate optical flow fields from $t-k$ frame to $t+k$ frame and from $t$ frame to $t+k$ frame. 
Our EDVF is more general than the original DVF.
Our EDVF network estimates optical flow fields $\hat{\bm v}_{t-k}^{t+k}$, $\hat{\bm v}_{t}^{t+k}$ and a weight map $\hat{\bm \omega}_{t+k}$ from frames ${\bm x}_{t-k}$ and ${\bm x}_{t}$ as shown in Fig.~\ref{fig:offline arch}.
Then, the predicted frame $\hat{\bm x}_{t+k}^{E}$ by the EDVF can be expressed as
\begin{eqnarray}
 \hat{\bm x}_{t+k}^{E} &=& \hat{\bm \omega}_{t+k} \otimes {\bm S}({\bm x}_{t}; \hat{\bm v}_{t}^{t+k} ) +
 (1-\hat{\bm \omega}_{t+k}) {\bm S}({\bm x}_{t-k}; \hat{\bm v}_{t-k}^{t+k} )\,,
\end{eqnarray}
where ${\bm S}({\bm x}; {\bm v})$ represents warping the frame ${\bm x}$ based on the optical flow field ${\bm v}$. 

Even if the EDVF can perfectly estimate the optical flow field, the EDVF cannot handle the occlusion and the appearance changes.
Therefore, in our prediction network, two refinement networks follow the EDVF.
Two refinement networks are the similar architecture of the hourglass-shaped network.
The inputs of the first refinement network are the observed frames ${\bm x}_{t-k}$ and ${\bm x}_{t}$, the estimated optical flow fields $\hat{\bm v}_{t-k}^{t+k}$ and $\hat{\bm v}_{t}^{t+k}$, and the estimated weight map $\hat{\bm \omega}_{t+k}$.
After the first refinment, the estimated frame $\hat{\bm x}_{t+k}^{R_1}$ can be expressed as
\begin{eqnarray}
 \hat{\bm x}_{t+k}^{R_1} &=& \hat{\bm x}_{t+k}^{E} + 
 {\bm g}_{R_1}({\bm x}_{t-k}, {\bm x}_{t}, \hat{\bm v}_{t-k}^{+k}, \hat{\bm v}_{t}^{t+k}, \hat{\bm \omega}_{t+k}; {\bm \theta}_{R_1} )
 \,,
\end{eqnarray}
where ${\bm g}_{R_1}$ represents the first refinement network, 
and
${\bm \theta}_{R_1}$ is the weights of the first refinement network.
The inputs of the second refinement network are the output of the EDVF and the refinement result by the first refinement network. 
The final predicted frame can be obtained by adding the second refinement component as
\begin{eqnarray}
 \hat{\bm x}_{t+k}^{R_2} &=& \hat{\bm x}_{t+k}^{R_1} + 
 {\bm g}_{R_2}(\hat{\bm x}_{t+k}^{E},\hat{\bm x}_{t+k}^{R_1}; {\bm \theta}_{R_2} )
 \,,
\end{eqnarray}
where ${\bm g}_{R_2}$ represents the first refinement network, 
and
${\bm \theta}_{R_2}$ is the weights of the first refinement network.

\subsection{Network Training}
\label{subsec:Network Training}
Before discussing the losses for the network training, we introduce an image quality measure. 
In the network-based restoration, a mean square error (MSE) in the pixel domain and in the gradient domain, a structural similarity (SSIM)~\cite{wang2004image}, and a perceptual similarity~\cite{zhang2018perceptual} are used for the loss function.
In this paper, we use the following image quality measure $\mu$ that consists of the above four metrics:
\begin{eqnarray}
 \mu(\hat{\bm x}, {\bm x}) &=& 
 \rho_{\rm MSEI} ||\hat{\bm x} - {\bm x}||_2^2
 + \rho_{\rm MSED} ||\nabla\hat{\bm x} - \nabla{\bm x}||_2^2
 \nonumber \\
 & &
 + \rho_{\rm SSIM} (1- {\rm SSIM}(\hat{\bm x}, {\bm x}) )
 + \rho_{\rm Per} ||\Phi(\hat{\bm x}) - \Phi({\bm x})||_1
 \,, \label{eq:mu}
\end{eqnarray}
where
${\bm x}$ is a ground truth image,
$\hat{\bm x}$ is an image to be evaluated,
$\nabla$ is a vector differential operator,
$||\cdot||_2$ represents L2-norm,
$||\cdot||_1$ represents L1-norm,
${\rm SSIM}$ is a function to evaluate a structure similarity~\cite{wang2004image},
$\Phi(\cdot)$ represents the Conv1-to-Conv5 of the AlexNet~\cite{krizhevsky2012imagenet},
and
$\{\rho_{\rm MSEI}, \rho_{\rm MSED}, \rho_{\rm SSIM}, \rho_{\rm Per}\}$ is a set of hyper-parameters.
We set $\{0.05, 0.001, 10, 10\}$ for the hyper-parameters of $\{\rho_{\rm MSEI}, \rho_{\rm MSED}, \rho_{\rm SSIM}, \rho_{\rm Per}\}$ in the section~\ref{subsec: Performance of Offline Network Architecture}, and $\{0.0001, 0, 10, 0\}$ for the hyper-parameters in the section~\ref{subsec:Future Frame Prediction on Online-Updating}.

The pre-trained prediction network is the prediction network, as shown in Fig.~\ref{fig:offline arch}, trained with the offline supervised manner in advance.
For the loss function for that training, we evaluate the image quality measures of internal predictions $\{ \hat{\bm x}_{t+k}^{E}, \hat{\bm x}_{t+k}^{R_1} \}$ and the final prediction $\hat{\bm x}_{t+k}^{R_2}$.
We also use the optical flow smoothness term assuming the optical flow is spatially smooth.
The loss function for the pre-trained prediction network is
\begin{eqnarray}
 L_{\rm Pre} &=& 
 \lambda_{E} \cdot \mu( \hat{\bm x}_{t+k}^{E}, {\bm x}_{t+k} )
 + \lambda_{R_1} \cdot \mu(\hat{\bm x}_{t+k}^{R_1}, {\bm x}_{t+k} )
 + \lambda_{R_2} \cdot \mu(\hat{\bm x}_{t+k}^{R_2}, {\bm x}_{t+k} )
 \nonumber \\
 & & 
 + \lambda_{\rm OF} ( ||\nabla\hat{\bm v}_{t-k}^{t+K}||_1 + ||\nabla\hat{\bm v}_{t}^{t+K}||_1 )
 \,,
\end{eqnarray}
where
$\mu$ represents the image quality measure in Eq.~\ref{eq:mu},
$\{ \hat{\bm v}_{t-k}^{t+K}, \hat{\bm v}_{t}^{t+K} \}$ is the optical flow fields estimated by the EDVF,
$\{\lambda_{E}, \lambda_{R_1}, \lambda_{R_2}, \lambda_{\rm OF} \}$ is a set of hyper-parameters.
In this paper, we set $\{ 2, 3, 7, 0.1 \}$ for the hyper-parameters of $\{\lambda_{E}, \lambda_{R_1}, \newline \lambda_{R_2}, \lambda_{\rm OF} \}$.
The pre-trained prediction network is randomly initialized, then trained with an end-to-end manner.

For online updating of the proposed framework as shown in Fig.~\ref{fig:onlinearch}, we use the following loss function: 
\begin{eqnarray}
 L_{\rm Ada} &=& \mu(\hat{\bm x}_{t}, {\bm x}_{t}) + \lambda_{C} \cdot \mu(\hat{\bm x}_{t}^{C}, {\bm x}_{t})
 \,,
\end{eqnarray}
where $\lambda_{C}$ is a hyper-parameter.
We set 0.1 for the hyper-parameter of $\lambda_{C}$.

Note that in the online update we predict $t$-frame from $(t-k)$- and $(t-2k)$- frames.
In the beginning, the weights of the continuous-updating prediction network are initialized by those of the pre-trained prediction network.
The continuous-updating prediction network and the weight estimation network are updated with an end-to-end manner, while the weights of the pre-trained prediction network are fixed in the test phase.

\section{Experiments}
\label{sec:Experiments}
This section presents the benchmark data set, evaluation metrics, and previous studies for comparison. Previously, the frame interval had been generalized to $k$. Hereafter, $k$ is set as $1$ to address the next-frame prediction problem.

\subsection{Experimental Environment}
\label{subsec:Experimental Environment}
We used the KITTI Flow data set~\cite{menze2015object} for offline experiments. This data set contains high-resolution scenes with large movements, and frequent occlusions. Therefore, it is challenging for future frame prediction problems. There are $200$ and $199$ videos for training and testing respectively, consisting of 21 frames. All frames are center cropped into 640$\times$320. The videos for training and testing were clipped in $3$ frames to focus on the $2$in-$1$out problem with $k=1$. Therefore, $3,800$ and $3,789$ samples were generated for training and testing respectively. 

For online-updating experiments, we used video clips ``2011\_09\_26\_drive\_0013'' and ``2011\_09\_26\_drive\_0014'' in KITTI RAW data~\cite{Geiger2013IJRR} with frame size 768$\times$256, and ``Stuttgart\_02'' in Cityscapes demo video~\cite{Cordts2016Cityscapes} with frame size 512$\times$256. Owing to the data variety, we used videos from Miraikan~\cite{Miraikan} and the imitation store in AIST~\cite{AIST}. All frames of Miraikan and AIST store were 960$\times$480.

We used Peak Signal-to-Noise Ratio (PSNR) and Structural Similarity Index Measure (SSIM)~\cite{wang2004image} metrics as performance evaluation of the methods. High PSNR and SSIM values indicate better performance. During SSIM and PSNR calculation, only 90\% of the center of the image was used. 

To compare with our pre-trained prediction network, existing high-performance network DVF~\cite{liu2017video} and DPG~\cite{Gao_2019_ICCV} were used.
We also employed PredNet~\cite{lotter2016deep} and ConvLSTM~\cite{xingjian2015convolutional}; however, we failed to tune those networks with several trials in different parameter settings.
DVF~\cite{liu2017video} and DPG~\cite{Gao_2019_ICCV} were selected as comparison methods because they exhibit state-of-the-art performances in frame prediction. We implemented the DPG by ourselves because of scarcity of public domain code. We adapted the public code of the DVF to our experimental environment. All networks, including the proposed network, were trained in Adam~\cite{kingma2014adam} with $lr=0.0001$ and $Epochs=100$ in offline training. In online training, networks were updated $1$ time by a set of three frames with a current input frame $x_t$ as ground truth and past frames $x_{t-1}$ and $x_{t-2}$ as input. In the experiment of online updating, data from KITTI, Cityscape, AIST store, and Miraikan were progressed with switching, and online updating of the data was performed simultaneously with performance evaluation. For performance evaluations of future frame predictions in online-updating, we used DVF, DPG, and our future frame prediction network for the ensemble network components. 

\subsection{ Performance of Future Frame Prediction Network Architecture}
\label{subsec: Performance of Offline Network Architecture}

\begin{wraptable}{r}{55mm}
 \caption{Comparison of pre-trained prediction on KITTI Flow data set}
  \begin{center}
   {\small
    \begin{tabular}{lll}
    \hline
    Method & SSIM \:\:\: & PSNR \\ \hline
    Repeat & 0.577 & 17.6 \\
    PredNet~\cite{lotter2016deep} & 0.162 & 9.58 \\
    ConvLSTM~\cite{xingjian2015convolutional} \qquad \qquad & 0.106 & 6.77 \\
    DVF~\cite{liu2017video} & 0.712 & 21.7 \\
    DPG~\cite{Gao_2019_ICCV} & 0.706 & 21.6 \\
    Ours & {\bf 0.718} & {\bf 22.1} \\
    \hline
    \end{tabular}
   }
  \end{center}
 \label{tab:kittiflow_entire}
\end{wraptable}

Table~\ref{tab:kittiflow_entire} summarizes the future frame prediction performances with the KITTI Flow data set.
From Table~\ref{tab:kittiflow_entire}, the our future frame prediction network performs the best in SSIM and PSNR. It achieved an SSIM of 0.718 and a PSNR of 22.1, whereas DPG~\cite{Gao_2019_ICCV} showed an SSIM of 0.706 and a PSNR of 21.6. The DVF~\cite{liu2017video} model achieved results inferior to those of our future frame prediction network and its SSIM and PSNR performances were comparable with those of DPG. The term ``Repeat'' means that the next frame is not predicted and current frame is responded as a prediction. Accordingly, a model showing a score higher than ``Repeat'' generates a significant prediction.

\begin{figure*}[t]
  \begin{center}
  \subfloat[GT]{
        \includegraphics[height=14mm]{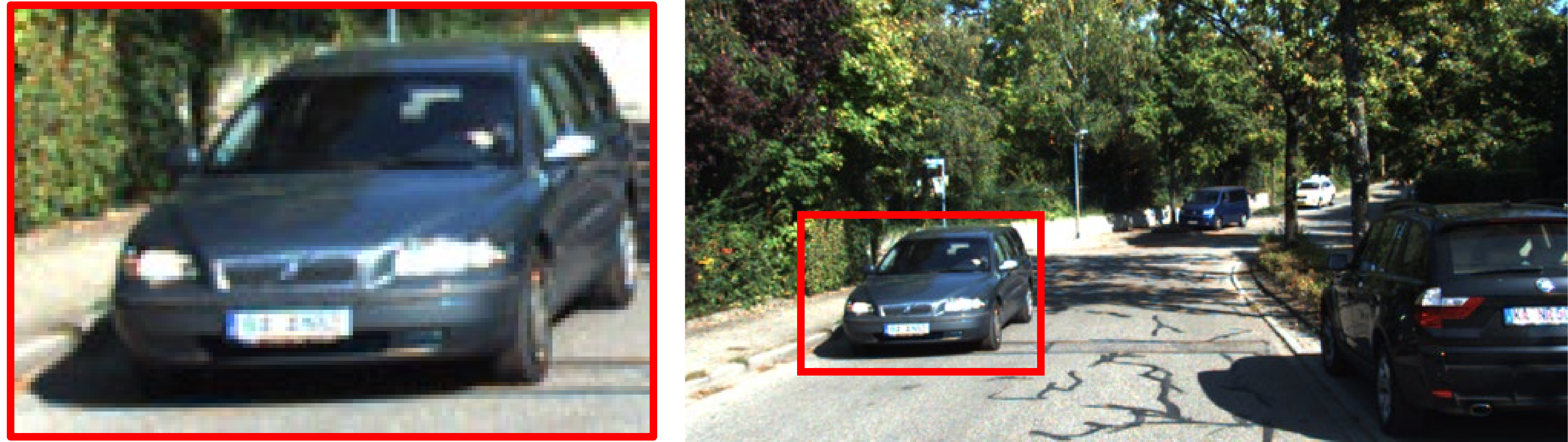}
        \includegraphics[height=14mm]{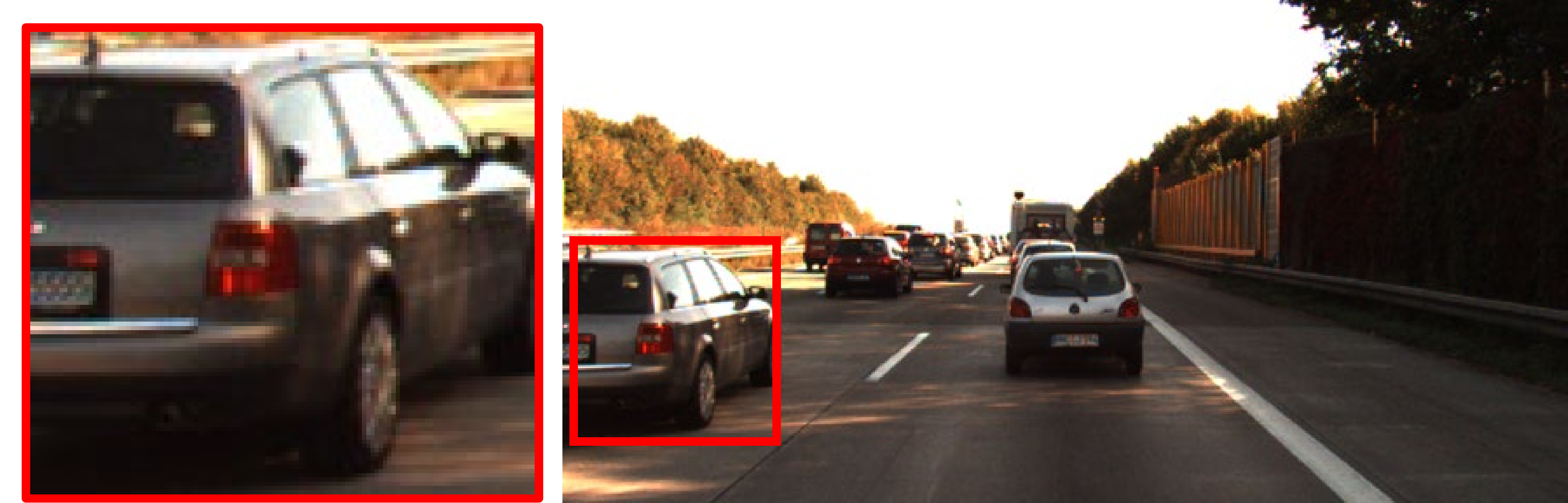}
        \label{fig:comparisonGT}
  }
  \\
  \subfloat[Repeat (current)]{
        \includegraphics[height=14mm]{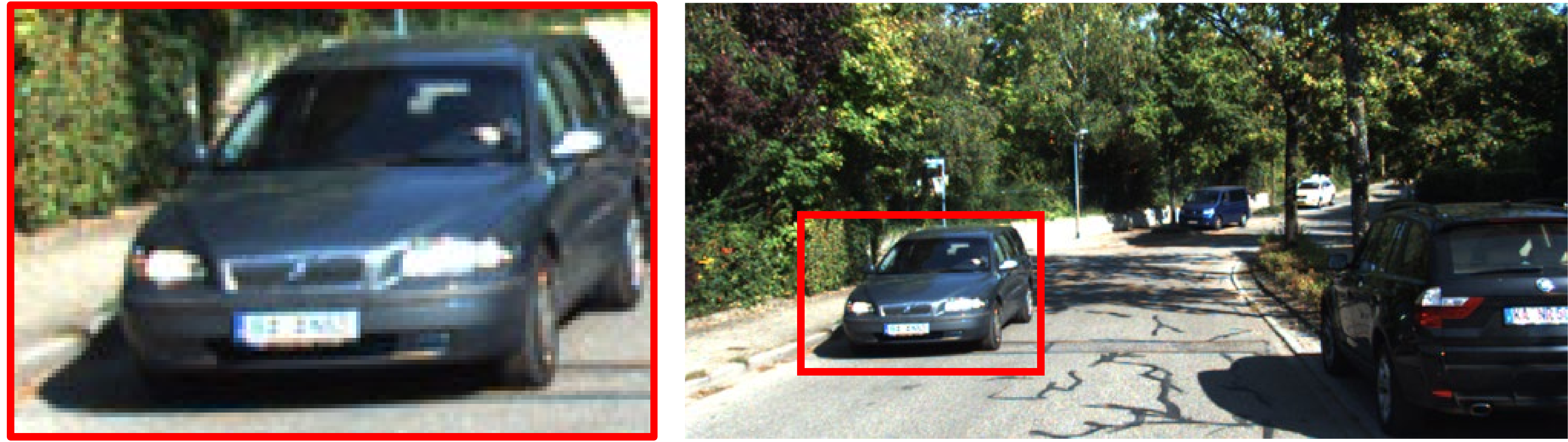}
        \includegraphics[height=14mm]{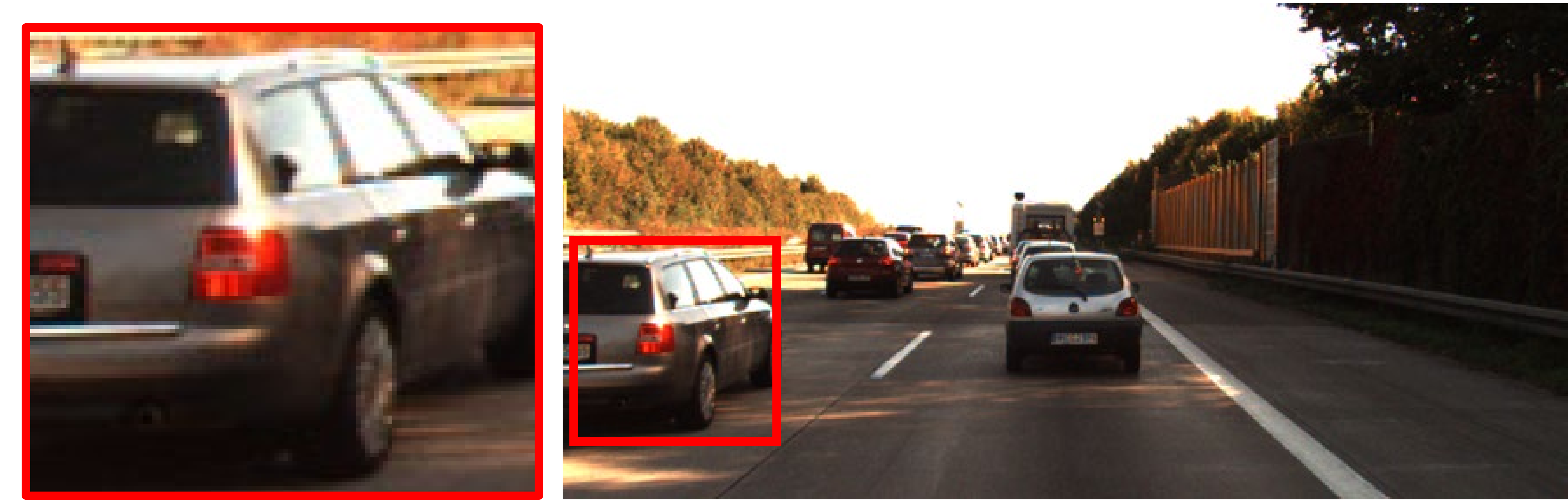}
        \label{fig:comparisonRP}
  }
  \\
  \subfloat[DVF]{
        \includegraphics[height=14mm]{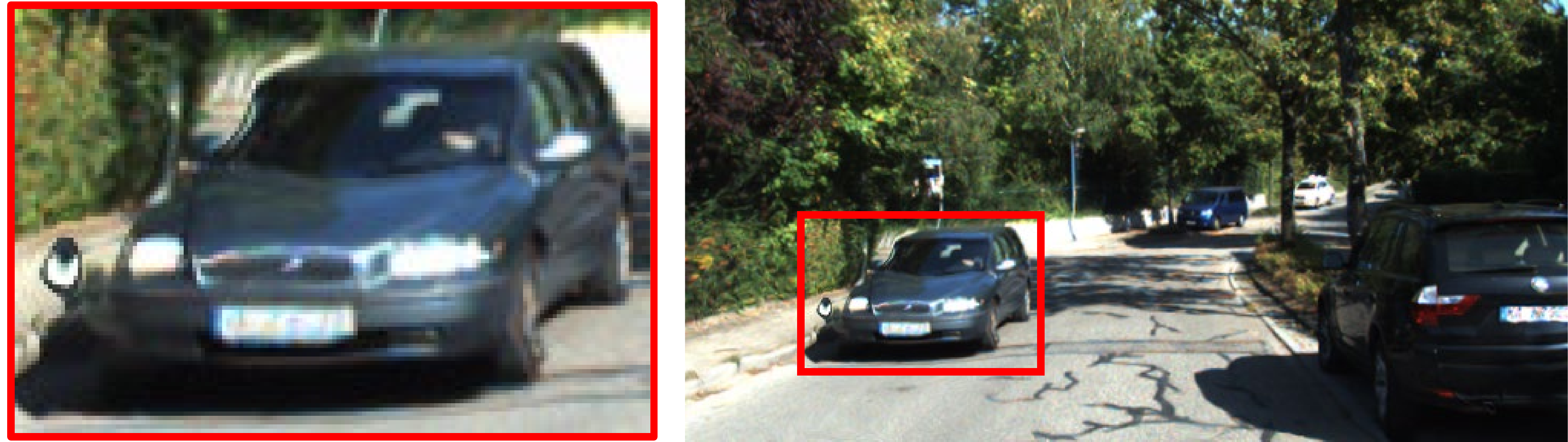}
        \includegraphics[height=14mm]{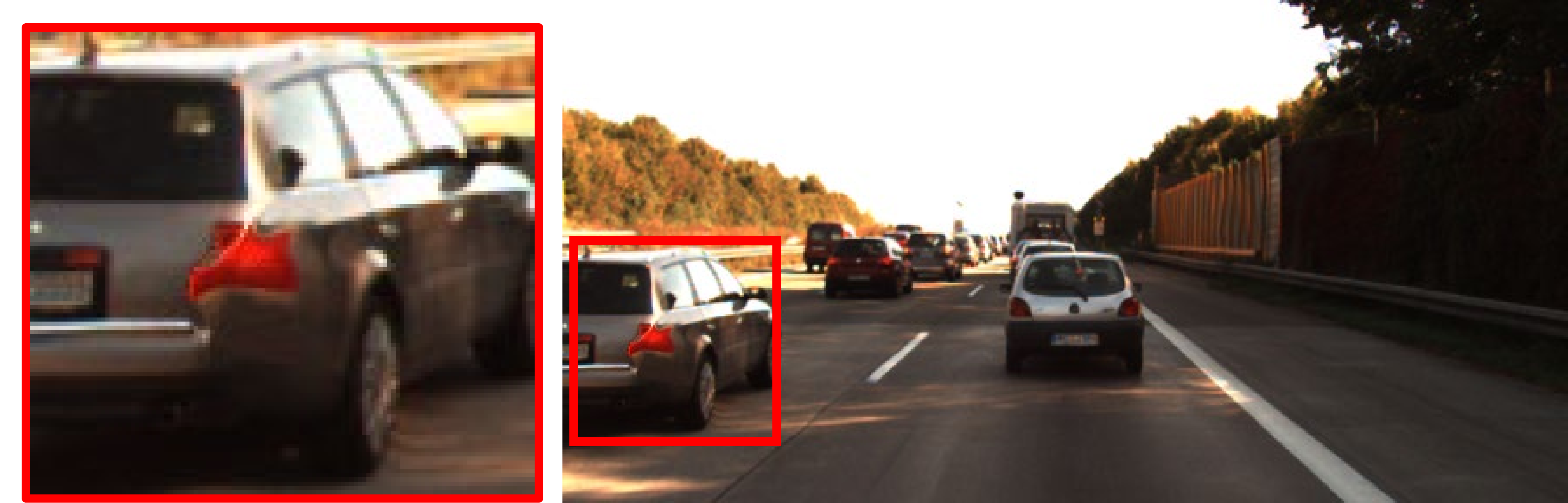}
        \label{fig:comparisonDVF}
  }
  \\
  \subfloat[DPG]{
        \includegraphics[height=14mm]{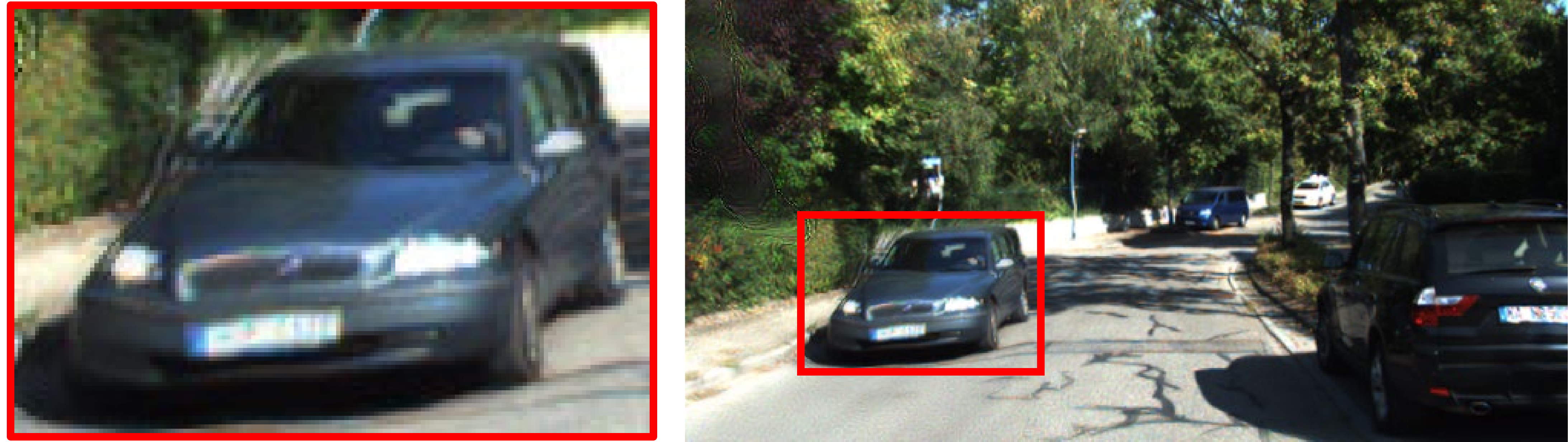}
        \includegraphics[height=14mm]{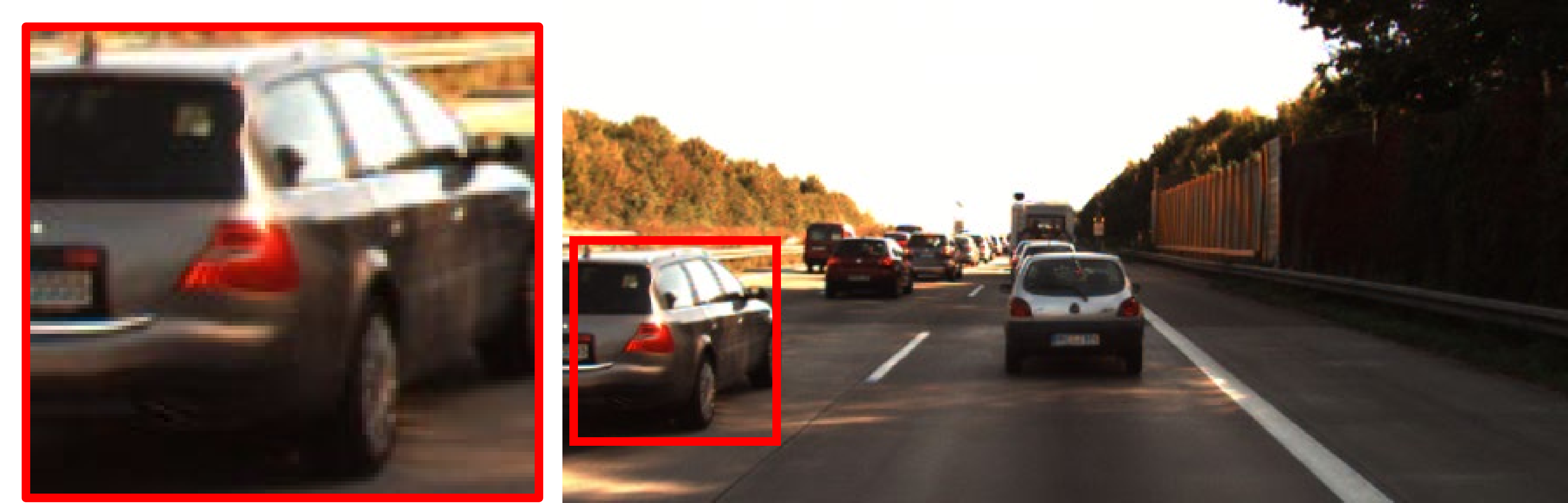}
        \label{fig:comparisonDPG}
  }
  \\
  \subfloat[Future frame prediction network (ours)]{
        \includegraphics[height=14mm]{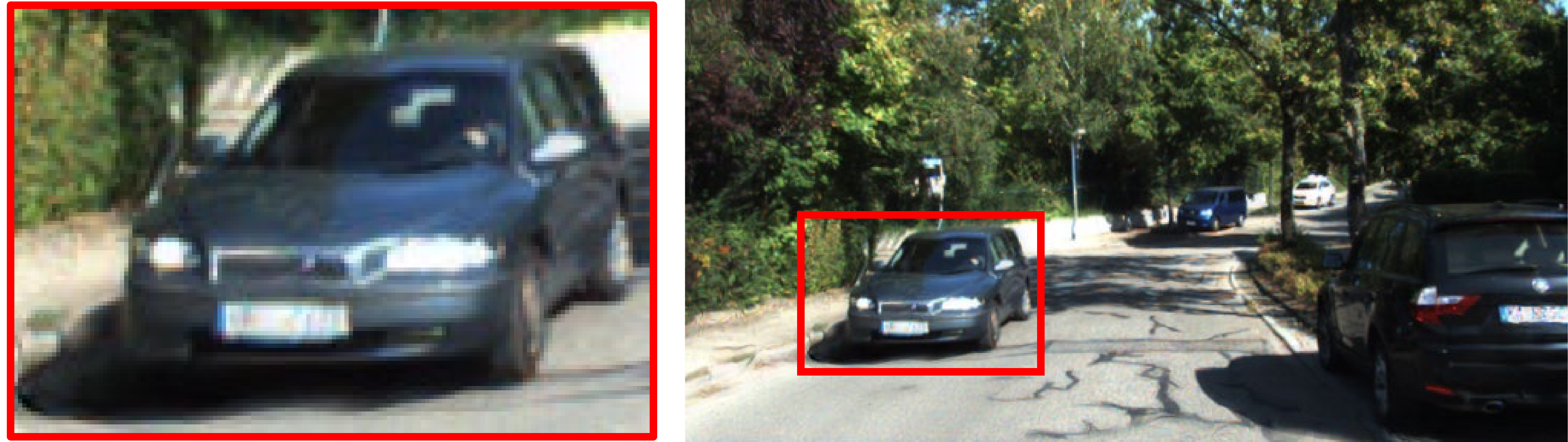}
        \includegraphics[height=14mm]{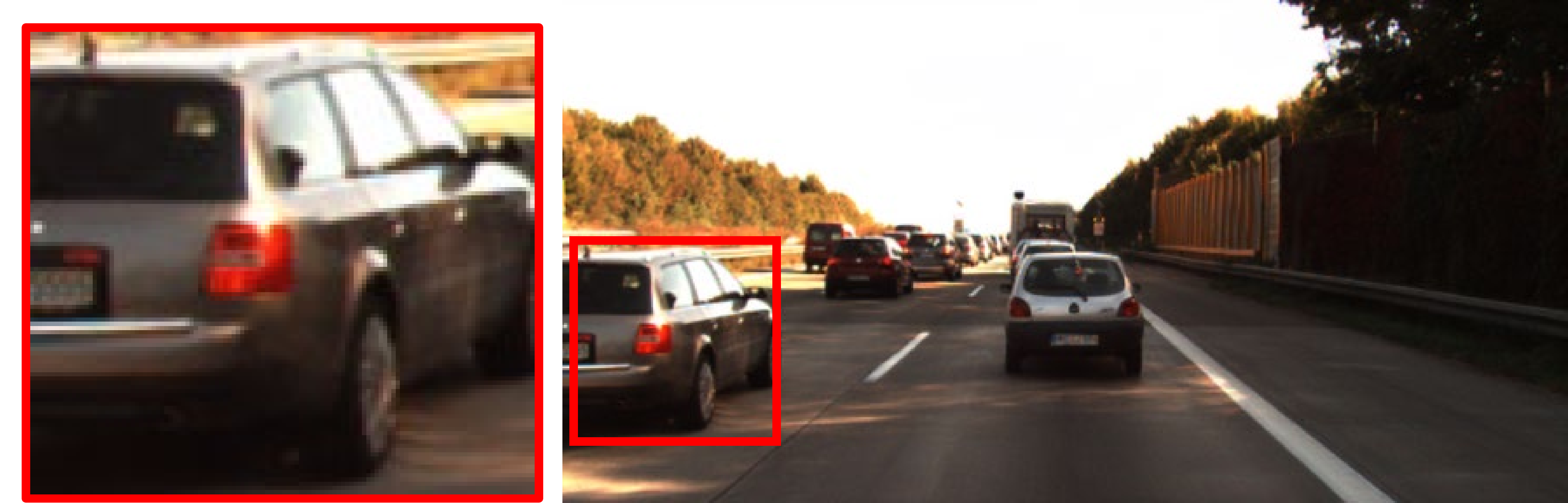}
        \label{fig:comparisonMRP}
  }
  \\
  \end{center}
  \caption{ Comparison of future frame prediction on the KITTI Flow data set.}
  \label{fig:offline result}
\end{figure*}

In Fig.~\ref{fig:offline result}, qualitative comparisons on KITTI Flow are depicted. All the frames predicted by DVF, DPG, and the our future frame prediction network are sharp and crisp. The our future frame prediction network predicted frames show minimal unnatural deformation, particularly in the red-highlighted region. However, the frames predicted by DVF and DPG include considerable unnatural deformation, particularly in the red-highlighted region. From the quantitative results in Table~\ref{tab:kittiflow_entire} and qualitative results in Fig.~\ref{fig:offline result}, the our future frame prediction network accurately predicted the flow and properly refined the future frame estimated by the predicted flow. More qualitative comparisons are shown in supplemental materials.

According to both quantitative and qualitative comparisons, we concluded that the future frame prediction network achieved the same level of performance as the existing state-of-the-art models.

\subsection{Future Frame Prediction on Online-Updating}
\label{subsec:Future Frame Prediction on Online-Updating}

\begin{figure*}[t]
  \begin{center}
  \subfloat[SSIM trends]{
        \includegraphics[width=100mm]{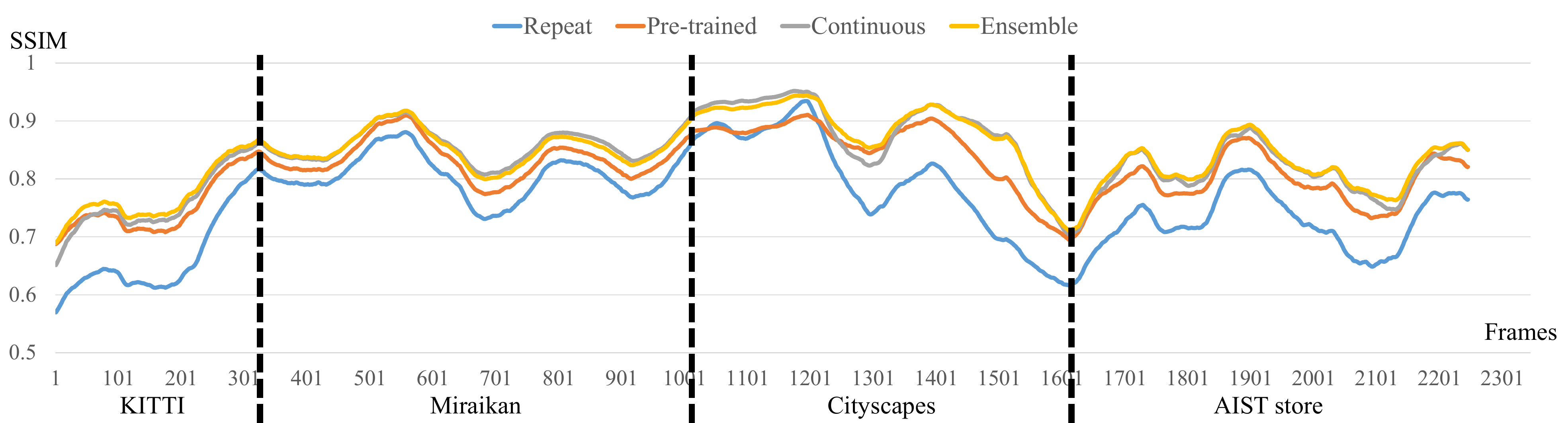}
        \label{fig:ssimtrend}
  } \\
  \subfloat[PSNR trends]{
        \includegraphics[width=100mm]{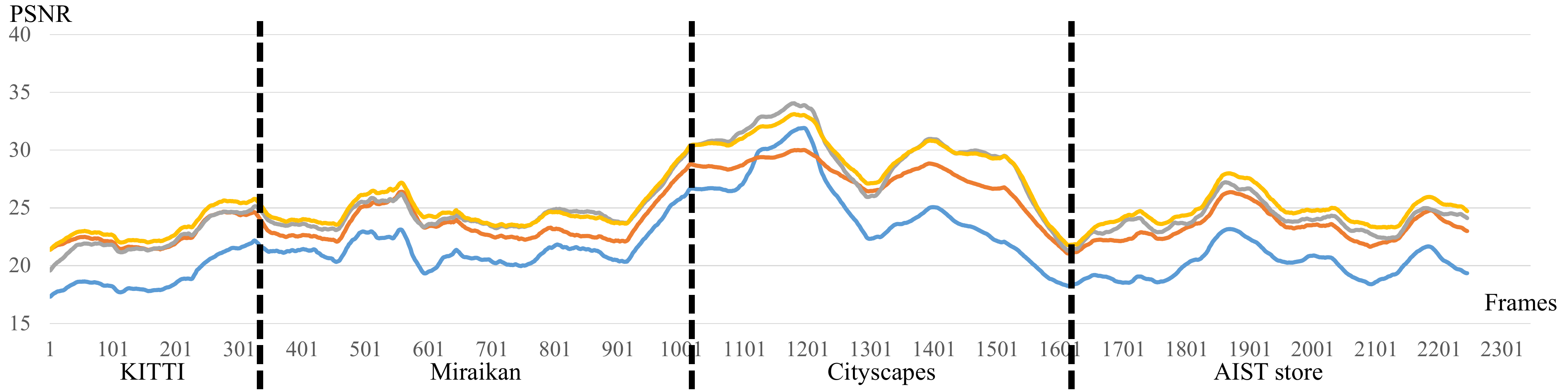}
        \label{fig:psnrtrend}
  } \\
  \end{center}
  \caption{ Tendencies of SSIM/PSNR in online updating. The data flow from left to right. Only the trends of the second round data flow are shown in here. Entire trends are shown in supplemental materials. \label{fig:online progress}}
\end{figure*} 

\begin{figure*}[t]
  \begin{center}
  \subfloat[GT]{
        \includegraphics[width=50mm]{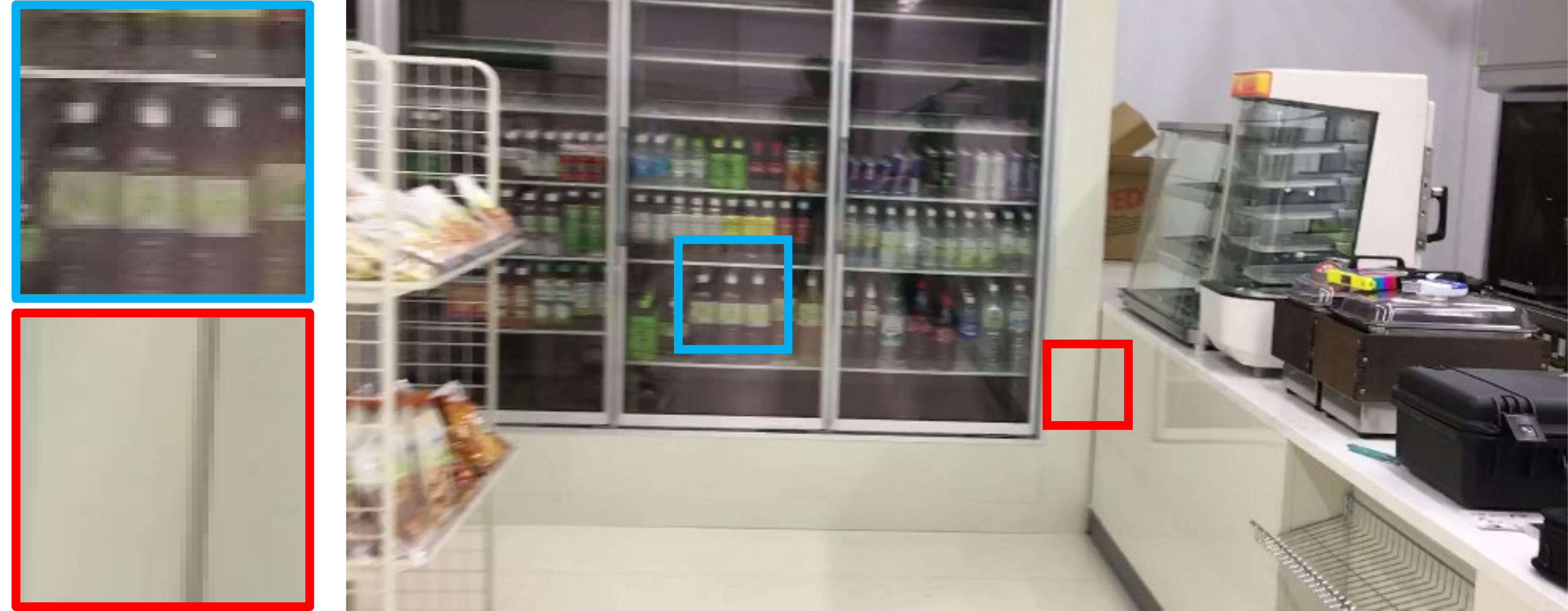}
        \includegraphics[width=50mm]{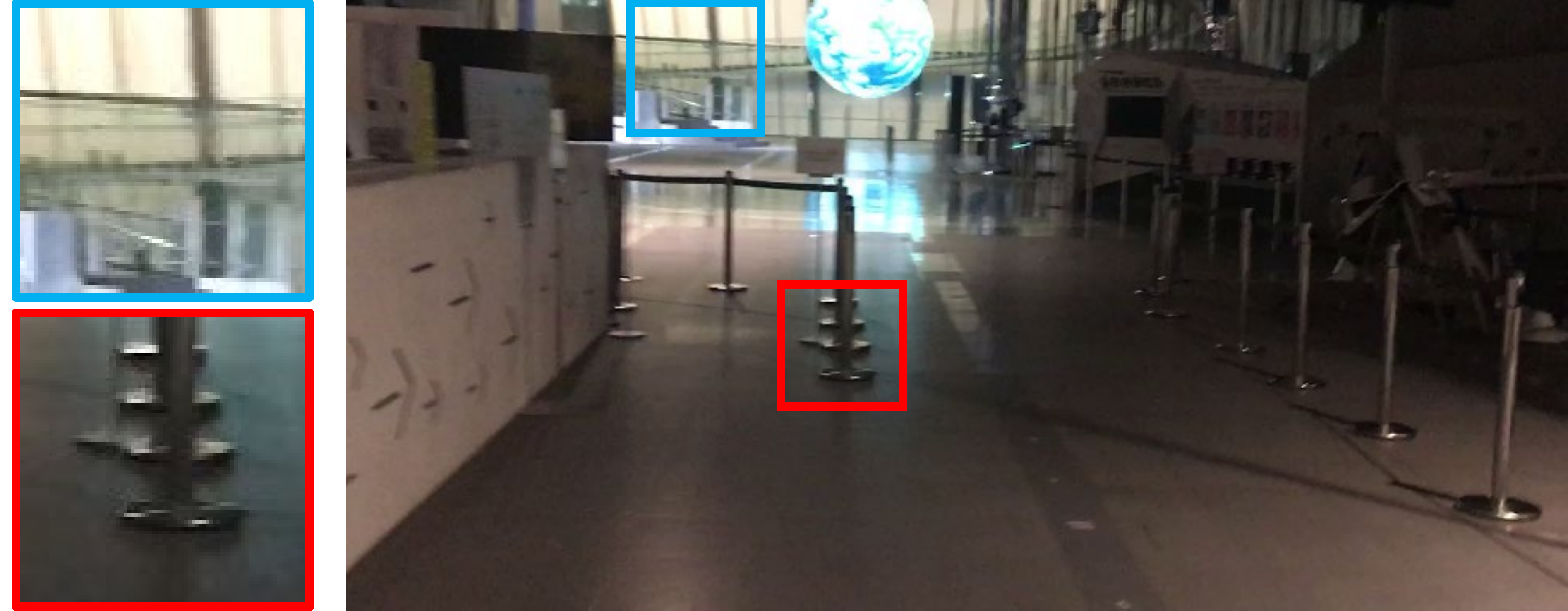}
        \label{fig:onlinecomparisonGT}
  }
  \\
  \subfloat[Pre-trained prediction network]{
        \includegraphics[width=50mm]{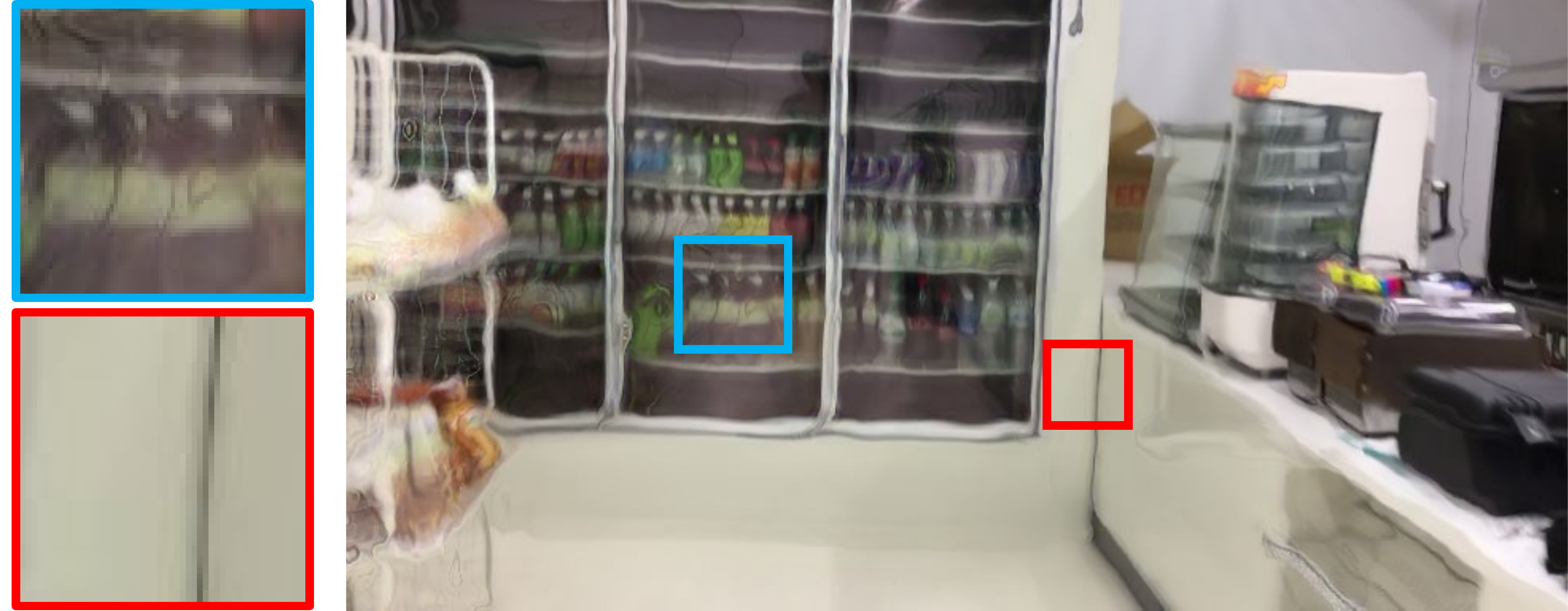}
        \includegraphics[width=50mm]{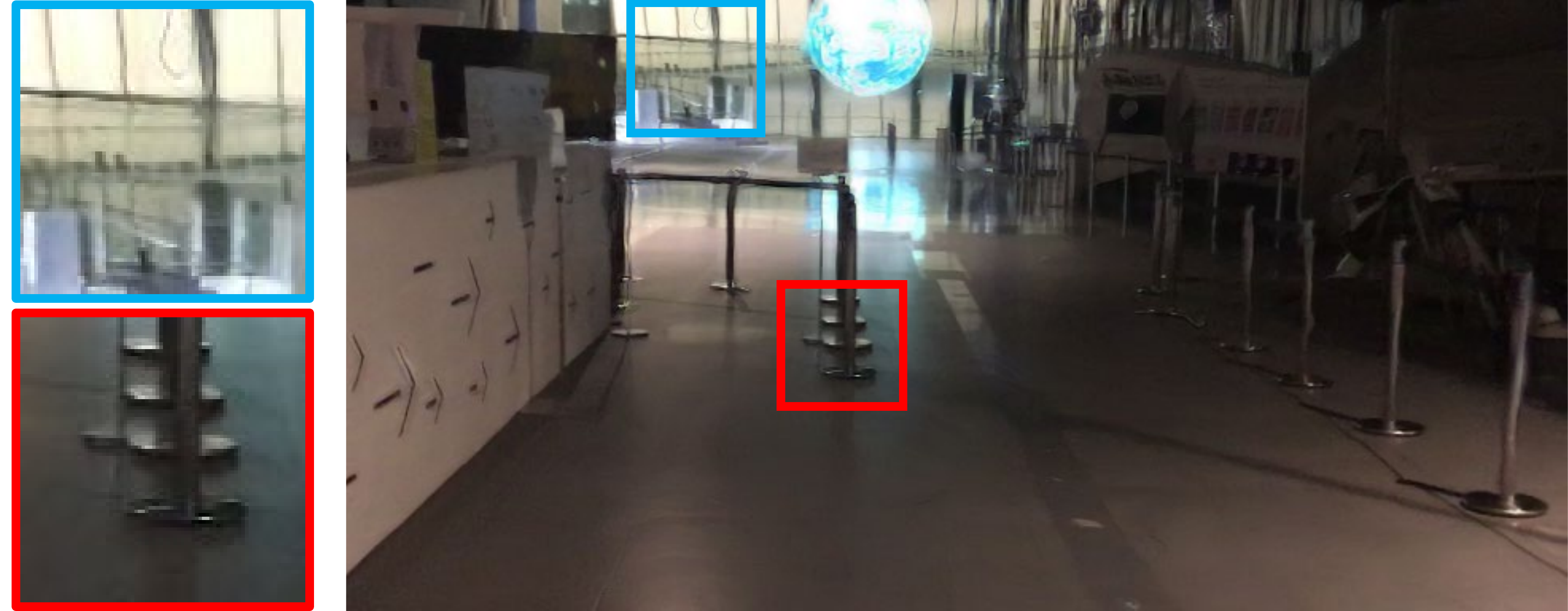}
        \label{fig:onlinecomparisonMRPNet}
  }
  \\
  \subfloat[Continuous-updating prediction network]{
        \includegraphics[width=50mm]{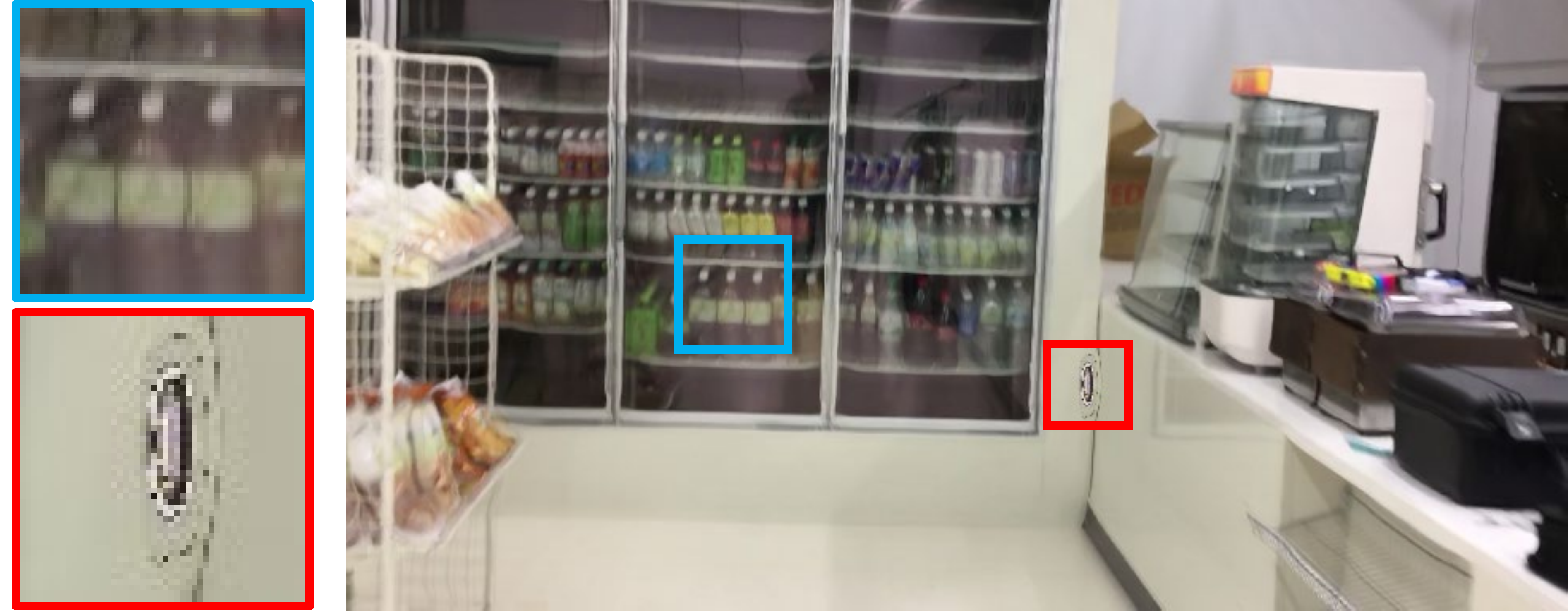}
        \includegraphics[width=50mm]{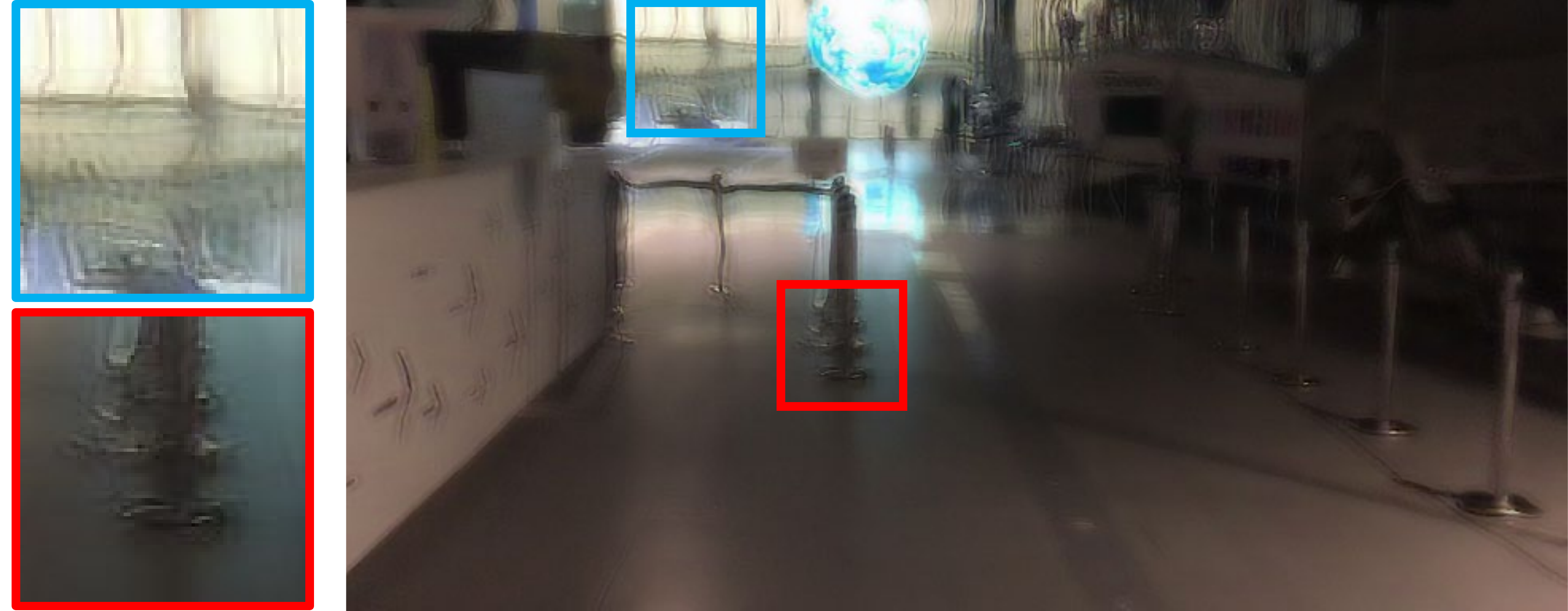}
        \label{fig:onlinecomparisonMRPNetUp}
  }
  \\
  \subfloat[Proposed ensemble network]{
        \includegraphics[width=50mm]{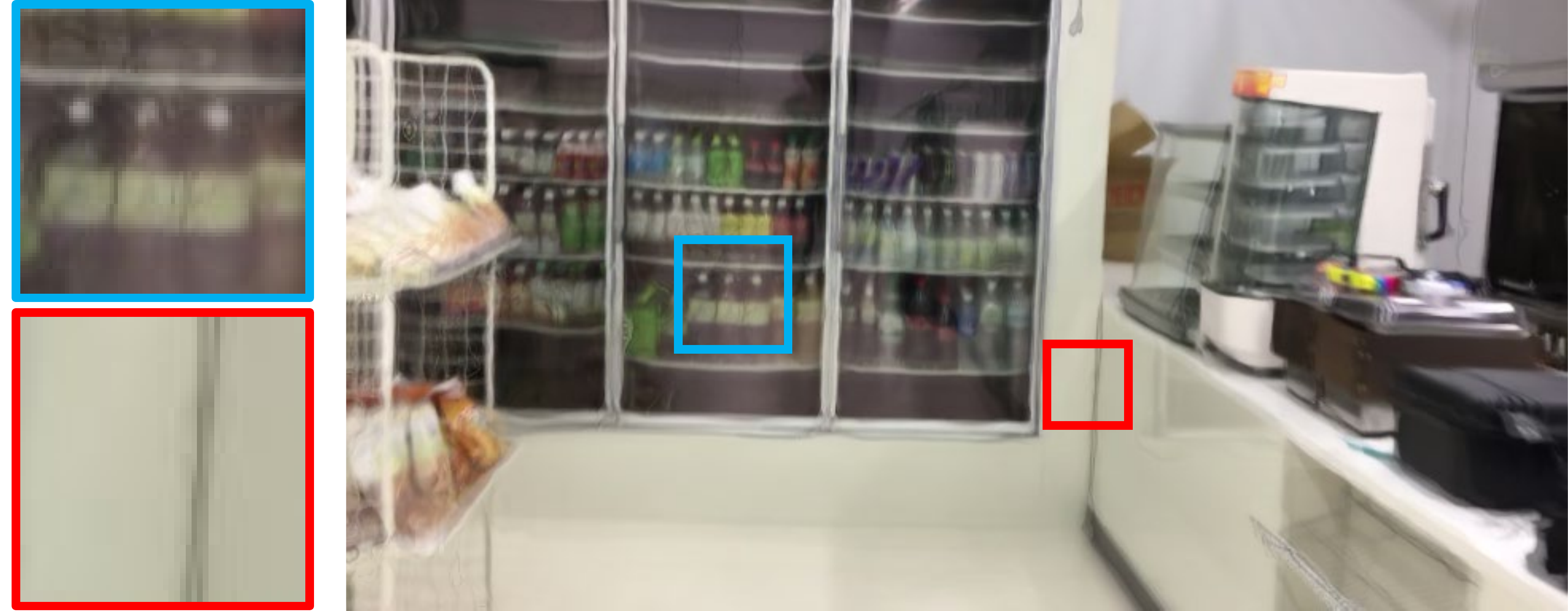}
        \includegraphics[width=50mm]{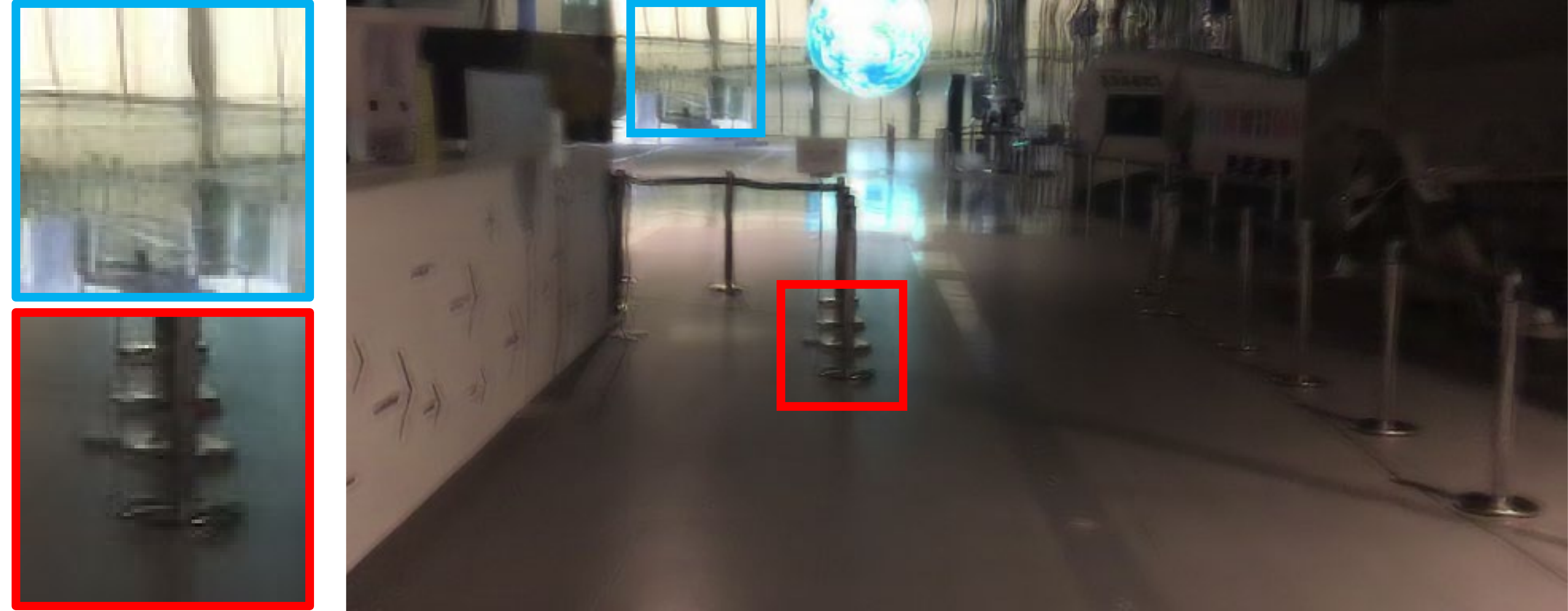}
        \label{fig:onlinecomparisonCPNet}
  }
  \\
  \end{center}
  \caption{ Examples of frame prediction for online updating evaluation. }
  \label{fig:online learning comparison}
\end{figure*} 

Table~\ref{tab:CPNet} shows the evaluation scores in the online-updating environment. The order of the scenes can be found in Fig.~\ref{fig:online progress}. The ``Ours", ``Pre-trained", ``Continuous-updating", and ``Ensemble" are meaning that proposed future frame prediction network, using the pre-trained prediction network only, using the continuous-updating prediction network only, and proposed ensemble network respectively. All networks in Table~\ref{tab:CPNet} use the network architectures and weights evaluated in Section~\ref{subsec: Performance of Offline Network Architecture}. From the scores of SSIM and PSNR in Table~\ref{tab:CPNet}, we can recognize that the proposed ensemble network using the future frame prediction network performs the best. No matter what network is used for the component, we can see that ``Ensemble'' achieves a higher score than ``Continuous-updating'', and ``Continuous-updating'' gets higher performance than ``Pre-trained''.

Figure.~\ref{fig:online progress} illustrates the SSIM and PSNR changes of ``Pre-trained'', ``Continuous-updating'', and ``Ensemble'' using the proposed future frame prediction network. A moving average was calculated on a window size 100 for better visualization. From Fig.~\ref{fig:online progress}-(a) and Fig.~\ref{fig:online progress}-(b), we can observe that there is a similar tendency between all manners. However, ``Ensemble'' shows higher metric values than other manners. There is a period where ``Pre-trained'' shows lower SSIM and PSNR than ``Repeat'' around the frame 1200. It means that the ``Pre-trained'' failed to predict in the period. ``Continuous'' produces higher performance than ``Ensemble'' on the frame 1200, but lower than ``Pre-trained'' on the frame 1300. It shows a characteristic of ``Continuous'', which has low stability. Overall, the graphs for ``Continuous'' and ``Ensemble'' are almost the same. This is because the difference both SSIM and PSNR is small, as described in Table~\ref{tab:CPNet}. However, "Ensemble" works to complement the part of "Continuous" that failed to predict, and visual examples are shown in Fig.~\ref{fig:online learning comparison}.

\begin{table}[t]
 \caption{Comparison of methods toward online updating environment}
  \begin{center}
   {\small
    \begin{tabular}{c|c|ccc|ccc|ccc}
    \hline
     & \multirow{2}{*}{Repeat} & \multicolumn{3}{c|}{Pre-trained} & \multicolumn{3}{c|}{Continuous-updating} & \multicolumn{3}{c}{Ensemble (Ours)} \\ \cline{3-11}
    & & DVF & DPG & Ours & DVF & DPG & Ours & DVF & DPG & Ours \\ \hline
    SSIM & 0.742 & 0.795 &\, 0.809 \,& 0.803 & 0.823 \, & 0.816 & 0.830 & 0.827 &\, 0.826 \,& {\bf 0.833} \\
    PSNR & 21.8 & 24.1 & 24.4 & 24.4 & 25.3 & 25.0 & 25.6 & 25.7 & 25.7 & {\bf 25.9} \\
    \hline
    \end{tabular}
   }
  \end{center}
 \label{tab:CPNet}
\end{table}

\begin{table}[t]
    \caption{Effectiveness of the ensemble network with updating in different frame intervals}
    \begin{center}
    {\small
    \begin{tabular}{l|cc}
    \hline
         Method & SSIM & PSNR  \\ \hline
         Our future frame prediction network (pre-trained) & 0.834 & 24.2 \\
         Ensemble network with updating every frames & 0.863 & 25.7 \\
         Ensemble network with updating every 3 frames & 0.856 & 25.5 \\
         Ensemble network with updating every 5 frames & 0.855 & 25.3 \\ \hline
    \end{tabular}
    }
    \end{center}
    \label{tab:async}
\end{table}

The visual comparison of ``Pre-trained'', ``Continuous-updating'', and ``Ensemble'' using our future frame prediction network is shown in Fig.~\ref{fig:online learning comparison}. In the left column, it can be seen that the prediction result of the ``Pre-trained prediction network'' is including unnatural deformation in the blue-highlighted region, and ``Continuous-updating prediction network'' is producing an aliasing-like error in the red-highlighted region. In the right column, we can see that the ``Pre-trained prediction network'' generates abnormal prediction, particularly in the blue-highlighted region. The ``Continuous-updating prediction network'' produces blurred prediction in both the blue- and red-highlighted region. However, problems appearing in the results of the ``Pre-trained prediction network" and ``Continuous-updating prediction network" are less observed in the ``Proposed ensemble network." More qualitative comparisons are shown in supplemental materials.

The quantitative and qualitative evaluations indicate that the proposed ensemble network is more stable and high-performing in unknown environments. In addition, the proposed pre-trained prediction network is suitable for the component of ensemble network. Consequently, proposed ensemble network which is a combination of the proposed methods, shows the most effective performance.

The effectiveness of the proposed network was verified through section~\ref{subsec: Performance of Offline Network Architecture} and~\ref{subsec:Future Frame Prediction on Online-Updating}. In the experiments, the frame interval is constant, but there are many variations in movement between frames. The employed data sets include all the walking, stopping, driving in the city, and driving on the highway. Therefore, when the overall performance is improved, the model can predict even in a large movement scene. In other words, the proposed networks keeps the effectiveness even if the frame interval is increased.

The proposed ensemble network with our future frame prediction network takes about 0.1 seconds to process KITTI RAW data~\cite{Geiger2013IJRR} with the GeForce GTX 1070. 
However, it takes about 0.3 seconds to update the proposed ensemble network with the same GPU. 
The proposed ensemble network can be implemented by updating in parallel with prediction toward data flow.
For example, the proposed network requires 0.1 seconds for KITTI RAW data~\cite{Geiger2013IJRR} to predict and 0.3 seconds to update. 
Because KITTI RAW data~\cite{Geiger2013IJRR} is 10fps, the network can predict all frames and update every three frames. 
Table~\ref{tab:async} shows the result of the parallel system at ``2011\_10\_03\_drive\_0047'' of KITTI RAW data~\cite{Geiger2013IJRR}. 
According to Table~\ref{tab:async}, although it is better to update with every new frame, we can confirm the effect of the proposed ensemble method even with updating every several frames.

\section{Conclusions}
\label{sec:Conclusions}

We proposed a novel ensemble network that accomplished state-of-the-art performance in future frame prediction in an online environment. The proposed ensemble network shows effectiveness toward unknown scenes that are not included in offline training data. From the experiments of online updating, we conclude that the design of the ensemble network is reasonable and effective.

We also proposed a future frame prediction network architecture used for the proposed ensemble network that achieved state-of-the-art performance in future frame prediction in an offline environment. The proposed network has improved the overall performance of ensemble network by providing pre-trained weight for initialization.

\bibliographystyle{splncs04}
\bibliography{egbib}
\end{document}